\def\year{2022}\relax
\documentclass[letterpaper]{article}
\usepackage{aaai22} 
\usepackage{times} 
\usepackage{helvet} 
\usepackage{courier} 
\usepackage[hyphens]{url} 
\usepackage{graphicx} 
\usepackage{graphicx}  
\usepackage{natbib}  
\usepackage{caption}  
\DeclareCaptionStyle{ruled}%
  {labelfont=normalfont,labelsep=colon,strut=off}
\title{Toward a New Science of Common Sense\footnote{This is a slightly revised version of a paper that appeared in the Proceedings of the Thirty-Sixth AAAI Conference on Artificial Intelligence (AAAI-22), 2022.}}
\author{
  Ronald J.\  Brachman,\textsuperscript{\rm 1}
  Hector J.\ Levesque\textsuperscript{\rm 2}
}
\affiliations {
    \textsuperscript{\rm 1} Jacobs Technion-Cornell Institute and
        Cornell University\\
    \textsuperscript{\rm 2} Dept.\ of Computer Science, University of
    Toronto\\[\smallskipamount]
    ron.brachman@cornell.edu,\,\, hector@cs.toronto.edu
}
\begin{document}
\maketitle
\begin{abstract}
Common sense has always been of interest in AI, but has rarely taken
center stage.  Despite its mention in one of John McCarthy's earliest
papers and years of work by dedicated researchers, arguably no AI
system with a serious amount of general common sense has ever emerged.
Why is that?  What's missing?  Examples of AI systems' failures of
common sense abound, and they point to AI's frequent focus on
expertise as the cause.  Those attempting to break the resulting brittleness
barrier, even in the context of modern deep learning, have tended to
invest their energy in large numbers of small bits of commonsense
knowledge.  While important, all the commonsense knowledge fragments in the world
don't add up to a system that actually demonstrates common sense in a
human-like way.  We advocate examining common sense from a broader
perspective than in the past.  Common sense should be considered in the 
context of a full cognitive system with history, goals, desires,
and drives, not just in isolated circumscribed examples. 
A fresh look is needed: common sense is worthy of its own 
dedicated scientific exploration.

\end{abstract}

\section{The Common Sense Gap}

The modern-era data-intensive machine learning juggernaut continues to
roll on, with a wide array of extraordinary results and significant
commercial impact.  But an increasing number of articles and books
\cite[for instance][]{Pavlus2020EasyQuestions,marcus2019rebooting}
point out that even the best of current AI falls short of the robust,
general intelligence envisioned by the field's founders.  Blunders made by
generally powerful systems have been
recounted, such as shocking misidentifications of
objects by otherwise accurate image recognition programs \cite{Szegedy2014,mitchell2019AI}.  Surprising gaffes of
seemingly remarkable systems like GPT \cite{vincent2020openAIGPT} have
been revealed as both humorous and
disturbing \cite{marcus-davisGPTBloviator}.  Self-driving cars make
terrifying unexplainable mistakes \cite{tesla-roadandtrack2021}.
Several authors
\cite[see, for example][]{levesque2017common,marcus2019rebooting,mitchell2019AI,toews2020GPTHype} have made the case
that AI is still missing something critical to avoiding these
mistakes, and they  identify the missing ingredient as what we
would normally call {\it common sense}.  But recent calls for a new
generation of post-modern AI systems with common sense give no real
prescription for getting there or a clear idea of what it would
really mean for an AI system to have it.  Our intention in this paper
is to stimulate the field into closing this critical gap.

\section{Expertise and the Brittleness Challenge}

Despite the name of the field, Artificial Intelligence's biggest
accomplishments have generally come from expertise rather than any
more general kind of intelligence.  Our greatest successes have been
on tasks in narrow domains or circumscribed challenge problems, such
as Go, facial recognition, infectious disease diagnosis, and the like.
The most obvious limit of this is what some have called
``brittleness''---the failure to produce reasonable outcomes (or, in many cases, any
outcomes at all) in the face of challenges just beyond the boundaries
of the expertise.  This was a well-known shortcoming of the 1980s wave
of expert systems, but as recent work has shown, it applies equally
well to systems trained with extensive amounts of
data \cite{Szegedy2014,marcus-davisGPTBloviator,GPT-3hype}.  AI systems
are often fragile and show a noticeable lack of common sense.  This
may not be a critical problem for a system that only plays chess and
whose entire world is limited to a chessboard and chess pieces.  But
AI's longer-term vision aspires to embed fully integrated systems in
the real world, where artificial agents will need to be able to cope
with a wide range of unanticipated events.

The emerging universe of self-driving cars provides a good example.
We expect such cars to operate on regular real-world streets with
natural phenomena occurring all around them---other drivers, signs and
signals, pedestrians and dogs, unpredictable weather and road
conditions, etc.  But we see that, at least for now, brittleness is
still rampant and current systems fail in ways that make it clear they
have no common sense to fall back on when their expertise meets its
limits \cite{tesla-roadandtrack2021}.  They make mistakes that seem
counterintuitive or just plain silly.  They cannot offer drivers
reasons for their behavior and we cannot correct them by offering
advice.  We end up with fragile, inscrutable, incorrigible systems
that can have serious and even fatal consequences when operating in
the real world---largely because they have no common sense.

\section{What Is Common Sense?}

If we want to develop a plan for building common sense into AI
systems, then the natural question to ask is, {\it what exactly is
it}?  The point of this paper is that that question has not been
sufficiently answered; if common sense were fully defined and its
technical challenges clearly articulated, there would be no need for a
new call to action.  Unfortunately much of the recent writing on
common sense in AI is not about what it is and how it can be realized,
but about its absence in current systems.  There are some thoughtful
treatises on how much is missing \cite[see, for
example][]{davis2015commonsense}, and there have been a number of
technical efforts focused on isolated fragments of {\it commonsense 
knowledge\/} (see below), but there is currently no real clarity on
how to build an AI system that consistently demonstrates common sense.

In our opinion, here is what common sense is about:
\begin{quote}\it
Common sense is the ability to make effective
use\\ of ordinary, everyday, experiential knowledge in\\ achieving
ordinary, practical goals.
\end{quote}
There is a lot to unpack in this characterization---words like ``ability,''
``effective,'' ``experiential,'' and ``practical'' are easy to gloss
over but each is intended quite specifically and is meaningful---but it is not our
intention to tackle that here.  (We take this up in considerable detail
in \cite{brachman-levesque2022MachinesLikeUs}.)  Rather, we want to
show how the idea of {\it making effective use of knowledge\/} 
leads to a set of scientific questions that we believe are
important for the field to consider in a systematic and unified
way. Progress on this front would have a very significant impact on
the ability of autonomous AI systems to operate in open-ended real
life.

\section{A Focus on Commonsense Knowledge}

Of course the consideration of common sense as an aspect of
intelligence is not a new phenomenon in AI.  Even John McCarthy's
earliest seminal paper in the field, ``Programs with Common Sense''
\cite{mccarthy1958programs}, mentioned it right in the title.
And a number of projects since then, including AI's longest
continuously running project---Cyc---have been said to have focused on
it \cite{lenat89building,MatuszekC2005Sfcs,Metz2016Cyc}.  But no
robust AI system generally and regularly exhibiting common sense has 
ever emerged from this line of
research.  Something critical is still lacking.

In our view, the problem is that almost all the attention on common
sense in AI to date has focused on the commonsense knowledge
that would be required, to the relative exclusion of several elements
that are key to the success of natural systems with common sense.
Researchers like Doug Lenat and others concentrated on the realm of
missing, tacit facts (like ``if someone dies, they stay dead'' or ``you
can't pick something up unless you're near it'') that most people
would know but that were never captured in formal knowledge bases built
through knowledge acquisition from experts.
Lenat published an interesting article that exemplified this, focusing on a number of
complex inferences about the play, {\it Romeo and Juliet\/}
\cite{Lenat2019Romeo}.  Missing ``obvious'' facts were
one reason that expert systems were stymied on edge cases, and their
pursuit was a well-justified avenue for addressing the brittleness
issue.  But it concentrated on only one part of a bigger problem,
which we will get to in a moment.  (In fairness, the Cyc project has also developed
a vast library of inference procedures that allow it to draw conclusions from
its knowledge---as of 2019 more than 1100 of these ``heuristic reasoning
modules'' were in place.  But while these help join together Cyc's millions of
bits of knowledge to produce novel conclusions, the system overall still
appears to be missing the bigger picture that we address below.)

Along similar lines,
researchers like Pat Hayes, Jerry Hobbs, Ernie Davis, Ray Reiter and
others developed formal theories of various aspects of
the commonsense world \cite{hayes1979naive,hayes1988second,hobbs1985formal,davis2014representations,reiter2001knowledge}.
The spectrum of areas of concern of this work was broad: naive physics,
time, plans, the minds of agents, even society and ethics.
The outputs of these efforts were generally foundational
axioms and rules; additional consequences of the specified axioms
were logically entailed.
To compute these consequences, standard inference mechanisms from logic
(monotonic or non-monotonic) were used.  Much of this work was sophisticated
and insightful, but was generally presented in the form of lists of axioms and proofs.
Some of it was implemented in systems that could draw interesting conclusions about
the world, but like Cyc, it generally did not result in integrated AI systems 
situated in the world trying to achieve goals.  Examples were mainly offered piecemeal
and in isolation.

A different view of commonsense knowledge was embodied in the now
decades-old memory-based efforts of Marvin Minsky and Roger Schank
and
colleagues \cite{minskyframes,schank-abelson-1977,schank-dynamic-memory1982}.
The emphasis there was more on the memories of past experiences than
on general truths about the world. The focus was less on deriving
conclusions from multiple facts and rules, and more on recognizing patterns and
drawing analogies between current circumstances and these remembered
experiences as a way of solving new problems.  Minsky's ideas about
frames and Schank's work on scripts, plans, and other memory
structures were often set in contrast with the more logical work noted
above, but in the end, this line of work also spent most of its energy
on knowledge and its organization.  It should also be noted that even
what we have called post-modern efforts in this space, like
COMET \cite{bosselut2019comet}, which look to build hybrid systems on top
of deep learning engines, are still focused on expanding 
knowledge bases.

\section{A Broader Perspective}

As mentioned, no system that can generally wield common sense in the 
frequent and effective way humans do has
ever emerged from these lines of work.  It has become increasingly
clear that no matter the scale of commonsense factual tidbits stored
in a knowledge base, this is not enough to get over the fundamental
hump of generally robust behavior outside the boundaries of expertise.

The crux of the issue is that {\it knowing even a vast number of
commonsense facts is simply not the same as having and exercising
common sense in the real world}.  When a person says to another, ``use
a little common sense here,'' they are not asking only to recall some
isolated bits of knowledge.  Having common sense is not the same as
being able to win some sort of obvious-fact trivia game 
(``Alex, what weighs more, a
wheelbarrow or a grizzly bear?'').  When we expect a person to use
common sense, what we are insisting on is the use of background
knowledge to influence what action to take or how to interpret an
unexpected experience.  Having common sense is substantially more than
having commonsense knowledge.  At the very least it is the appropriate
and timely {\it application\/} of this knowledge that is critical.

One gets the sense from many presentations of commonsense knowledge in
action in AI papers that we are creating the equivalent of locally-scoped
``fact calculators,'' which can be fed a number of piecemeal facts and
(if all goes well) will spit out inferences.  This is
clearest in the case of systems based directly on logic ({\it \`a la}
McCarthy/Cyc): you give the inference engine some axioms, push the
``INFER''  button, and 
it can come to some conclusions, perhaps even interesting or unexpected
ones.  But it's working in isolation of any contextual
situation, goals, and prior history; those
are all in the head of the user pressing the buttons.
The conclusions are not of any interest or utility to the calculator itself.  The {\it Romeo and
Juliet\/} example mentioned above is just like this: the author supplies the question
and circumscribes what he wants the system to do with it.  The system is not
an agent, working in the context of its own history or goals, and is not
using its conclusions to decide what it wants to do next.  It's just a fancy calculator.  Yes, it is using commonsense
{\it knowledge}---but it's not {\it acting commonsensically}.

The more memory-based research ({\it \`a la} Minsky/Schank) comes
closer to advocating the use of knowledge in context.  The suggestions
in a frame for what to look at in a scene, or what other knowledge
structures to look at to explain a phenomenon, can direct reasoning in
a more contextually-relevant way. But these are only first steps to
more broadly intelligent behavior.  What is missing is a global
architecture that invokes the incremental reasoning steps of a frame
system at the right times with focus on the right issues and controls
its application to solve the problem at hand.  How does an agent
decide which chunks of knowledge to look at next, how they should
combine with other chunks of knowledge, when to go back for another
try, and even when to give up and try something different?

Even efforts that purport to focus on commonsense reasoning rather
than purely on commonsense knowledge seem to have this kind of
``isolated steps'' feel.  Take efforts in qualitative reasoning, for
example.  The inference mechanics in such efforts expressly attempt to
avoid getting bogged down in mathematical detail, thereby reflecting
what seems to be a common human trait of quick, qualitative analysis.
This is no doubt important and will play a role in future AI systems
with common sense.  But what is the bigger picture here?  How can this
kind of stepwise computation be used at the right times and in the
right ways by an agent being confronted with an unexpected event in the
world?  How might a qualitative inference capability be integrated
with a more logical one, to assure an integrated agent's survival and
practical success in the world?  Being able to reason quickly and
qualitatively is crucial, but it doesn't in itself give you common sense
(even with a huge knowledge base of tiny obvious rules).  One might say
that this is commonsense reasoning in the small, but not commonsensical
thinking in the large.

Related to this, and generally missing in AI systems, are the
circumstances and methods of invocation of common sense.  In humans,
common sense is not always active.  Much of what humans do every
day is routine: our days are dominated by mindless, rote behavior.  We
follow set patterns that we've learned over time---brushing our teeth,
walking the dog, even driving to work on an uneventful day.  What
causes us to break out of a routine and think more deliberately about
what we are doing? In our view, when something unusual happens during
the execution of a mindless routine, the first resort is common
sense---using past experience to quickly and plausibly explain the
unusual event or to guide the next action to take. This seems to be
one of the primary things common sense is for.  If common sense fails
to provide a plausible next action or if its suggestion fails, a more
thoughtful, deeper analysis can then take over.  The speed and
facility of common sense spares cognitive workload and often provides
adequate solutions to problems and guidance for actions.  But when and
how it is invoked, and how it decides what background knowledge to
use, are questions that have not been addressed by prior AI
efforts.

\section{Time to Consider a New Science}

In our opinion, what is needed here is a new start on an old problem.
Common sense has generally not been treated in AI as a first-class
problem, studied from start to finish as a whole.  As a field we need
to step back from building larger training sets or capturing millions
of explicit commonsense tidbits, and analyze common sense in and of
itself as a significant facet of intelligence.  Then we can approach
its implementation in AI systems as an integrated whole.  What we need
is a new science of common sense.

What would such a new science aspire to cover?
\begin{itemize}

\item {\bf Commonsense knowledge}: Years of work on large 
commonsense rule bases like Cyc will not be wasted, although in our
view, much more focus needs to be placed on the {\it experiential\/} basis of
common sense.  How are experiences remembered, generalized, and
organized so that they can be called to mind when needed?  Thought
should be given to mechanisms for representing baseline ontological
information, general rules of thumb, exceptions, and a host of other
items that distinguish commonsense knowledge from other forms of
knowledge.  Many agree that there are some key domains that
need to be accounted for, like knowledge of the physical world,
understanding of other agents, time, causality and some others.
The Minsky/Schank lines of thinking should be reexamined
and their complementarity to the more logic-based McCarthy/Cyc line
should be investigated.

\item {\bf Commonsense reasoning}: This needs a careful
analysis, definition, and prescription for implementation.  For any
chain of commonsense inference, we need to be clear on just what the
inputs and the expected outputs are going to be.  Are all logical
consequences going to be computed? If so, what is the plan to ensure
this can be done quickly enough?  If not, what exactly is going to be
left out?  Rapid, plausibly sound inference seems to characterize
common sense but has been underdeveloped in AI.  Analogy and similarity seem
central to common sense and their study should be reinvigorated.
Both commonsense reasoning in the small---for example, mechanisms
for drawing plausible conclusions from inexact statements or dealing with
exceptions---as well as broader
commonsensical thinking mechanisms---like rough planning and
quick-and-dirty decision-making---have important roles.

\item
{\bf Cognitive architecture}: Critical to the overall phenomenon of
common sense is when and how it is invoked and how it fits with the
rest of cognition, perception, and action, and how metareasoning
comes into play.  How it interrelates with goals and drives and
overall priorities will be important.  Key questions related to focus
of attention will need to be addressed, including how attention is
focused on relevant items of background knowledge, and how it moves
away from one thing to a more promising one.  We'll also need to sort
out mechanisms that smoothly allow the agent to give up on commonsense
reasoning and move to a more analytical, heavier-workload reasoning
effort as needed.

\item

{\bf Learning}: It is generally agreed that the bulk of the basis for
common sense in humans is learned from experience.  Machine learning
is the dominant technical thread in AI right now, but it has
focused heavily on classifiers and predictive
technologies like transformers.  What would a learning machine look
like if it were targeted to learning general knowledge of the sort one
sees in Cyc?  How would a machine go about learning how to use any
knowledge it may have already learned?  Can some of the architectural
considerations mentioned above be learned or must they be innate in
the underlying framework of an AI system?  Finally, can common sense
itself be {\it taught\/} after the fact?  There are some self-help
books out there that seem to imply that it can; if so, what are the
implications for AI systems?

\item 
{\bf Explanation and advice-taking}: We believe that no
system that purports to be autonomous should be deployed without
common sense---how common sense relates to autonomy, explanation, and
advice-taking will need to be part of this new endeavor.  Autonomous
systems need to be responsible for their actions and need to be open
to taking advice as necessary from others.

\end{itemize}
These are the key areas that come to mind immediately.  
There are no doubt other major areas of investigation that the field will
prioritize when it starts to grapple with common sense as a whole.

\section{Building on Prior Work}

A wide variety of great work that has been done in the past will surely
need to be incorporated into our new undertaking.  
There is no doubt that the commonsense knowledge crafted over
many years in projects like Cyc and the others mentioned above will be of value.  Minsky's
frame ideas tantalizingly hinted at how the knowledge structures
should be used, for things like ``differential diagnosis'' and
reconceptualizing.  Schank's work on scripts and reminding will play
an important role.  And there are other sources of insight from many corners of AI that can be
drawn upon, only a sample of which we have mentioned here.

While the psychology literature is surprisingly short on
analyses of common sense in adult humans, the work of Sternberg and
colleagues on what they call ``Practical Intelligence'' is clearly
relevant \cite{sternberg2000practical}.  (Sternberg cites four modes of intelligence, and equates
one of these---practical intelligence---exactly with common sense.)  From an AI
perspective there are limitations in this work's perspective, but it
is worth integrating into the big picture of synthetic common sense.
Along a different dimension, psychologists have posited a ``cognitive
continuum,'' in which it is postulated that common sense fits in its
own position between ``intuition'' and ``analysis'' (see \cite{hammond1987direct}; Hammond calls
common sense ``quasi-rationality'').  This kind of account
may inspire how to build an integrated AI system that allows common
sense to be used at the right time and to show its value in frequent
avoidance of heavy cognitive burdens.
Along related lines, the psychologist Daniel Kahneman postulates a distinction between
rapid intuitive processing in what he calls ``System 1'' and more thoughtful, methodical
reasoning in what he calls ``System 2'' \cite{Kahneman-FastandSlow-2011}.  It is not clear
at this stage how well Kahneman's classification aligns with the common sense/expertise distinction
we believe is central to AI. At the very least, it appears that common sense as we see it
does not fall neatly within System 1 or within System 2; it instead shares
some characteristics with each, and has some critical features not
accounted for in either. 

Given its focus on simple reasoning using models rather than general rules and abstractions,
the work of psychologist Philip
Johnson-Laird is worth taking into account \cite{johnson1983mental}.  
We would also need to account for the difference between common sense and the
broader notion of {\it rationality}, and explore and build on relationships
to bounded and minimal rationality \cite{simon1990bounded,cherniak1986minimal}.
The work of
Gary Klein on intuitive decision-making is also of potential use \cite{klein2007power}.  And
prior psychology work on prototypes and exemplars
is worthy of incorporation \cite{rosch-cognition1978,smith-medin1981categories}.
From an AI perspective, a
number of prior efforts on cognitive architecture \cite{kotseruba202040}
will be relevant as well.  Classical AI efforts on qualitative reasoning and naive physics will also
play a role in the expanded research endeavor we envision.


If all goes well, an ongoing government-sponsored program expressly aimed at common sense should uncover insights and mechanisms to be folded into our bigger picture:
over the last several years, DARPA has run a program on ``Machine Common Sense'' \cite{DARPAMCSProgram}.  One part of the program looks to construct a knowledge repository capable of answering queries about commonsense phenomena, a thrust that seems to be of a kind with the work focused on commonsense knowledge mentioned above.  This effort should produce some valuable resources for the community, but will likely manifest the same kind of context-independent behavior about which we expressed concern.  However, another thrust of the program focuses on learning from experience, and will investigate early human development in search of inspiration for the development of a foundation for machine common sense. 
Within this program, a major project by cognitive scientist Joshua Tenenbaum and others has revealed some important insights and technology approaches by attempting to reverse-engineer the core of human common sense \cite{Tenenbaum-BBS2017,Tenenbaum-GECCO2021}. Tenenbaum points out that humans have ``more content than we thought'' in their starting state and ``more [learning] mechanisms than we thought, some of them very smart.''  These insights and the project's outlining of a ``commonsense core'' will no doubt play an important role in a new science with common sense as its focus.

Recent work on Project Mosaic at the Allen Institute for Artificial Intelligence is also worth consulting \cite{AIAI-ProjectMosaic-benchmarks-website}.  Among the items of interest in this project is its consideration of benchmarks for measuring progress in machine common sense.  Evaluation frameworks and methods are important in any science, and one that focuses on common sense is no different.

\section{Core Research Questions}

Taking common sense seriously as its own integrated subject matter
leads to a number of important research questions.  The key questions
of the field will need to be articulated.  Here are some candidates:

\begin{itemize}

\item What exactly is common sense?  What technical definition best suits 
the needs of AI?

\item What are appropriate tests for the presence of common
sense?  How can we tell if we are getting closer to building it into our
AI systems?


\item 
How is experiential knowledge represented, accessed, and brought to
bear on current situations?  What is the role of analogy?  How does
the ability to recognize something or see something as another thing
(or even as an instance of an abstract concept) develop and get used?

\item 
How is commonsense knowledge learned as new experiences happen?  How
is the update different when knowledge is acquired through language?

\item 
What ontological frameworks are critical to build into an AI system?
Are there special properties of the knowledge of the physical world that need to be handled 
in a way that is different from its non-physical counterparts? 

\item What is the relationship between common sense and the broader notion
of rationality (including bounded rationality, minimal rationality, etc.)?

\item 
What overall architecture is best suited for the multiple roles of
common sense?  What mechanism(s) should be used to invoke common sense
out of routine, rote processing, and then to sometimes go beyond it to
more specialized forms of expertise?

\item What role, if any, does
metareasoning play?
\end{itemize}

\section{Refocusing on Common Sense as a Phenomenon}

Since the beginning of AI, McCarthy and a limited cohort of
researchers have set their sights on giving computers common sense.
Unfortunately, while the last sixty-five years has provided us impressive
technology that works on narrow problems, it has failed to deliver the
ability to deal with the unpredictable open world: we do not have AI
systems that can use common sense to solve life's rampant mundane
problems and respond reasonably and practically to unforeseen events.
It is frequently said of AI that it can rival the most expert of human
experts in many fields but cannot do the everyday things that a
six-year-old can.  Our belief is that the field's thinking about
common sense has been limited and has unwittingly cornered itself into a focus on
commonsense knowledge and isolated islands of inference, and has never looked
into what common sense as a whole may be.  We need to move from
systems with large amounts of independent knowledge fragments to
systems that show that they can use common sense in their everyday
interactions with the world.  The way to do this is to step back and
consider common sense in all its glory, including not just the
knowledge equivalent of sound bites, but how it is based in experience
and how and when it is applied.  To get to true AI---systems that can
be deployed and operate autonomously in the real world---we need to
tackle common sense head on, as a first-class subject of study.

\bibliography{newaaai22}
\end{document}